\documentclass[runningheads]{./styles/llncs}

\usepackage{graphicx}
\usepackage{booktabs}
\usepackage{trees}
\usepackage{shortcuts}
\usepackage{subcaption}
\usepackage{caption}
\newcommand{\ra}[1]{\renewcommand{\arraystretch}{#1}}
\newcommand*\circled[1]{\tikz[baseline=(char.base)]{
            \node[shape=circle,draw,inner sep=.5pt] (char) {\scriptsize{#1}};}}

\begin{document}

\title{IMS at the PolEval 2018: \\ A Bulky Ensemble Dependency Parser meets 12~Simple Rules for Predicting Enhanced Dependencies in Polish}

\titlerunning{IMS at the PolEval 2018}

\author{Agnieszka Falenska \and Anders Bj\"{o}rkelund \and Xiang Yu \and Jonas Kuhn}

\institute{Institute for Natural Language Processing , University of Stuttgart, Germany
\email{\{falenska,anders,xiangyu,jonas\}@ims.uni-stuttgart.de}}

\maketitle

\begin{abstract}
This paper presents the IMS contribution to \polevalFull{}.\footnote{\url{http://poleval.pl}} We submitted systems for both of the Subtasks of Task~1.
In Subtask (A), which was about dependency parsing, we used our ensemble system from \conllstFull{}. The system first preprocesses the sentences with a CRF POS/morphological tagger and predicts supertags with a neural tagger. Then, it employs multiple instances of three different parsers and merges their outputs by applying blending.  The system achieved the second place out of four participating teams. In this paper we show which components of the system were the most responsible for its final performance.

The goal of Subtask (B) was to predict enhanced graphs. Our approach consisted of two steps: parsing the sentences with our ensemble system from Subtask (A), and applying 12 simple rules to obtain the final dependency graphs. The rules introduce additional enhanced arcs only for tokens with ``conj'' heads (conjuncts). They do not predict semantic relations at all. The system ranked first out of three participating teams. In this paper we show examples of rules we designed and analyze the relation between the quality of automatically parsed trees and the accuracy of the enhanced graphs. 


\end{abstract}

\section{Introduction}

This paper presents the IMS contribution to \polevalFull{} (\poleval{}). The Shared Task consisted of three Tasks: (1) Dependency Parsing, (2) Named Entity Recognition, and (3) Language Models. Our team took part only in the Task (1) and submitted systems for both of its Subtasks (A) and~(B).

The goal of the Subtask (A) was predicting morphosyntactic analyses and dependency trees for given sentences. The IMS submission was based on our ensemble system from \conllstFull{} \cite{udst:overview}. The system (described in detail in \cite{bjorkelund-EtAl:2017:CONLL} and henceforth referred to as \imsoriginal{}) relies on established techniques for improving accuracy of dependency parsers. It performs its own preprocessing with a CRF tagger, incorporates supertags into the feature model of a dependency parser \cite{ouchi-duh-matsumoto:2014:EACL2014-SP}, and combines multiple parsers through blending (also known as reparsing; \cite{sagae-lavie:2006:HLT-NAACL06-Short}).

The original system only needed few modifications to be applied in the PolEval18-ST setting. First, the organizers provided gold-standard tokenization so we excluded the tokenization modules from the system. Second, one of the metrics used in the PolEval18-ST was BLEX. While the metric takes lemmas into consideration we added a lemmatizer to the preprocessing steps. Finally, \imsoriginal{} was designed to run on the TIRA platform \cite{tira}, where only a limited amount of CPU time was available to parse a multitude of test sets. 
The maximal number of instances of individual parsers thus had to be limited to ensure that parsing would end within the given time. Since in the \poleval{} setting the parsing time was not limited we removed the time constraint from the search procedure of the system. We call the modified version \imsnew{}.

The aim of Subtask (B) was to predict enhanced dependency graphs and additional semantic labels.
Our approach consisted of two steps: parsing the sentences to surface dependency trees with our system from Subtask (A), and applying a rule-based system to extend the trees with enhanced arcs. Since the \poleval{} data contains enhanced dependencies only for conjuncts, our set of manually designed rules is small and introduces new relations 
only for tokens with ``conj'' heads (it does not predict semantic labels at all). 

All components of both submitted systems (including POS tagger, morphological analyzers, and lemmatizer) were trained only on the training treebank. Out of all the additional resources allowed by the organizers we  used only the pre-trained word embeddings prepared for \conllstFull{}.\footnote{\url{https://lindat.mff.cuni.cz/repository/xmlui/handle/11234/1-1989}} We did not employ any Polish-specific tools
as they (or the data their models were trained on) was not among the resources allowed by the organizers.

The remainder of this paper is organized as follows. Section~\ref{sec:taskA} 
discusses our submission to Subtask (A) and analyzes which components of the system were the most responsible for its final performance. In Section~\ref{sec:taskB} we describe our submission to Subtask (B), show examples of the designed rules, and analyze the relation between the quality of automatically parsed trees and the accuracy of the enhanced graphs. Our official test set results are shown in Section~\ref{sec:test} and Section~\ref{sec:conclude} concludes.

\section{Subtask (A): Morphosyntactic prediction of dependency trees}
\label{sec:taskA}

The focus of Subtask (A) was morphosyntactic prediction and dependency parsing. The training and development data contained information about gold-standard tokenization, universal part-of-speech tags (UPOS), Polish-specific tags (XPOS), universal morphological features (UFeats), lemmas, and dependency trees. The dependency trees were annotated with the Universal Dependencies (UD) \cite{ud} according to the guidelines of UD v. 2.\footnote{\url{http://universaldependencies.org/}} To make the Shared Task more accessible to participants, the test data was released with baseline predictions for all preprocessing steps using the baseline \udpipe{} 1.2 system \cite{udpipe:2016}.

\subsection{System description}
\label{sec:description}

\begin{figure}[t]
\includegraphics[width=\textwidth]{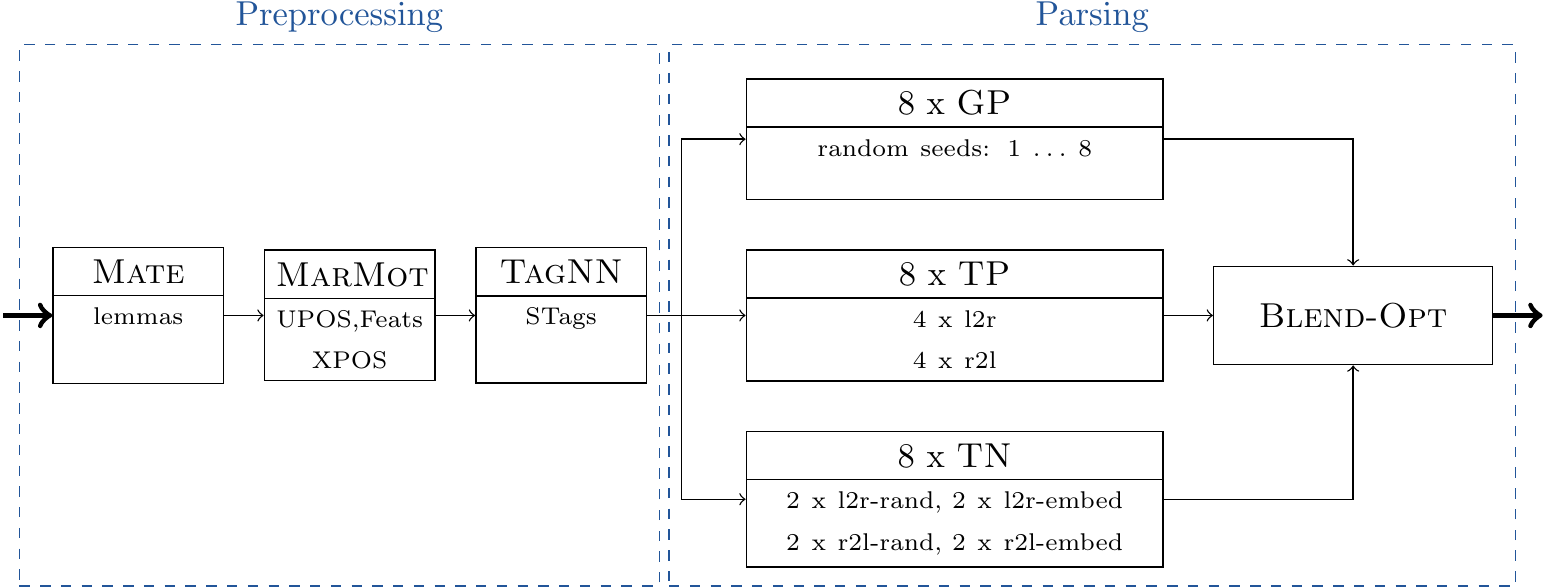}
\caption{\imsnew{} system architecture.} 
\label{fig:architecture}
\end{figure}

Figure~\ref{fig:architecture} shows an overview of the \imsnew{} system architecture. 
The architecture can be divided into two steps: preprocessing and parsing. The system uses its own preprocessing tools, so we did not utilize the baseline \udpipe{} predictions provided by the ST organizers. All the preprocessing tools annotate the training data via 5-fold jackknifing. 
The parsing step consists of running multiple instances of three different baseline parsers and combining them into an ensemble system. All the trained models for both of the steps, as well as code developed during this Shared Task will be made available on the first author's page.

Below we give a summary of all the components of the system and describe changes introduced to the \imsoriginal{} system needed to adapt it to the \poleval{} setting. \\

\noindent \textbf{Lemmatization}
is not performed by \imsoriginal{}. Since BLEX, one of the metrics used in the PolEval18-ST, takes lemmas into consideration, we added a lemmatizer to the preprocessing steps. For this purpose we used lemmatizer from the \texttt{mate-tools}  with default hyperparameters.\footnote{\url{https://code.google.com/archive/p/mate-tools/}} \\

\noindent \textbf{Part-of-Speech and Morphological Tagging}
is performed within \imsoriginal{} by \marmot{}, a morphological CRF tagger \cite{mueller2013}.\footnote{\url{http://cistern.cis.lmu.de/marmot/}} UPOS and UFeats are predicted jointly. Since \imsoriginal{} did not use XPOS tags, we added an additional CRF tagger predicting only XPOS tags (separately from other preprocessing steps). We used \marmot{} with default hyperparameters. \\

\noindent \textbf{Supertags}
\cite{Joshi:1994} are labels for tokens which encode syntactic information, e.g., the head direction or the subcategorization frame. \imsoriginal{} follows \cite{ouchi-duh-matsumoto:2014:EACL2014-SP} and extracts supertags from the training treebank. Then, it incorporates them into the feature models of all baseline dependency parsers. Supertags are predicted with an in-house neural-based tagger (\xiangtag{}) \cite{Yu:2017}.\footnote{\url{https://github.com/EggplantElf/sclem2017-tagger}} \\

\noindent \textbf{Baseline parsers}
used by \imsoriginal{} differ in terms of architecture and employed training methods. The system uses three baseline parsers: 
(1) The graph-based perceptron parser from \texttt{mate-tools} \cite{bohnet:2010:PAPERS}, henceforth referred to as \textbf{\mate{}} (the parser has been slightly modified to handle features based on supertags and shuffle training instances between epochs).\footnote{Since there are no time constraints in the \poleval{} (unlike the CoNLL 2017 Shared Task), \mate{} is applied to all sentences, cf. \cite{bjorkelund-EtAl:2017:CONLL} for details on how some sentences were skipped to save time in the \imsoriginal{} system.} (2) An in-house transition-based beam-perceptron parser \cite{bjorkelund-nivre:2015:IWPT}, henceforth referred to as \textbf{\abt{}}. (3) An in-house  transition-based greedy neural parser \cite{yu-vu:2017:ACL}, henceforth referred to as \textbf{\xiang{}}. 
We use the default hyperparameters during training and testing of all the three baseline parsers. \\

\noindent \textbf{Blending,}
i.e., combining  outputs of multiple different baseline parsers, can lead to improved performance \cite{sagae-lavie:2006:HLT-NAACL06-Short}. \imsoriginal{} parses every sentence with each baseline parser and combines all the predicted trees into one graph. It assigns scores to arcs depending on how frequent they are in the predicted trees. Then it uses the Chu-Liu-Edmonds algorithm \cite{chu-liu:1965,edmonds:1967} to find the maximum spanning tree in the combined graph. For every resulting arc it selects the most frequent label across all the labels previously assigned to it.

To enlarge the number of parsers taking part in the final ensemble 
\imsoriginal{} trains multiple instances of each baseline parser using different random seeds:
(1) eight \mate{} instances, (2) eight \abt{} instances which
differ in the direction of parsing -- four parse from left to right (\abt{}-l2r) and four from right to left (\abt{}-r2l), (3) eight \xiang{} instances which differ in the direction of parsing and the used word embeddings  -- four use pre-trained embeddings (\xiang{}-l2r-embed, \xiang{}-r2l-embed) and four use randomly initialized embeddings (\xiang{}-l2r-rand, \xiang{}-r2l-rand).

The final component of the \imsoriginal{}  system (\blend{}) selects the best possible blending setting. It checks all the possible combinations  of the above-mentioned instances ($9 \times 5 \times 5 \times 3 \times 3 \times 3 \times 3 = 18,225$ possibilities) and selects the one which achieves the highest LAS score on the development set. 
The original \imsoriginal{} limits the maximal number of instances of individual parsers to ensure that parsing will end within a restricted time. 
Since in the \poleval{} setting the parsing time was not limited we removed the time constraint from the search procedure \blend{}.

Finally, since the UD guidelines do not allow multiple root nodes, we re-attach all excessive root dependents in a chain manner, i.e., every root dependent is attached to the previous one.

\subsection{Evaluation of the components of the system }

In this section we evaluate all the components of the submitted \imsnew{} system with the evaluation script provided by the ST organizers. We use the \udpipe{} 1.2 system (as provided by the ST organizers) as a baseline through all the steps. \\

\noindent \textbf{Preprocessing and Supertags.}
We begin with evaluating the preprocessing components of our system on the development data (see Table~\ref{tab:preprocess}). We find that \udpipe{} is much better at predicting lemmas than \texttt{mate-tools} and it surpasses it by more than 10 points. On the contrary, \marmot{} outperforms \udpipe{} on all the other tagging tasks, with the highest gain of more than two points on the task of predicting morphological features.

\begin{table}[t]
\ra{1.3}
\centering
	\begin{tabular}{lccccc} \toprule
		& & Lemma & UPOS & XPOS & UFeats \\ \midrule
		UDPipe & & 94.41 & 97.24 & 86.50 & 88.30 \\
		IMS & & 84.09 & 97.69 & 87.00 & 90.52  \\ \bottomrule
	\end{tabular}

\caption{Preprocessing accuracy (F1 score) on the development set.}
\label{tab:preprocess}
\end{table}
    
To see how the above-mentioned differences influence the parsing accuracy we run the baseline parsers (\mate{}, \abt{}, and \xiang{}) in four incremental settings: (1) using UPOS and morphological features predicted by \udpipe{}, (2) replacing UPOS and morphological features with \marmot{}'s predictions, (3) adding lemmas, (4) adding supertags. Table~\ref{tab:gains} shows LAS scores for the three baseline parsers  for the consecutive experiments. Replacing \udpipe{}'s UPOS and morphological features with the predictions from \marmot{} improves accuracy by 0.42 points on average. The introduction of lemmas improves only the \mate{} parser and leads to minuscule improvements for the other two. The step which influences the final accuracy the most is the addition of supertags. It brings an additional 0.9 points on average (with the biggest gain for \abt{} of 1.54 points). \\

\begin{table}[t]
\ra{1.3}
\centering
	\begin{tabular}{lp{.3cm}cp{.3cm}ccc} \toprule
		& & \udpipe{} &  & \marmot{} & +lemma & +STags \\ \midrule
		GP & &  83.36  & & +0.27 & +0.30 & +1.03  \\
		TP (l2r) & & 81.80 &  & +0.55 & +0.01 & +1.54 \\
		TN (l2r-rand) &  & 82.77 &  & +0.43 & +0.03 & +0.15  \\ \midrule
		\textit{average} & & 82.64 &  & +0.42 & +0.11 & +0.90 \\ \bottomrule
	\end{tabular}
	\caption{Gains in parsing accuracy (LAS) for by incrementally replacing the \udpipe{} preprocessing baseline.}
\label{tab:gains}
\end{table} 

\noindent \textbf{Parsing and Blending.}
Table~\ref{tab:devparse} shows parsing results on the development set.
The relation between baseline parsers (rows \circled{2}, \circled{3}, and \circled{4}) is the same as in \cite{bjorkelund-EtAl:2017:CONLL}: \mate{} is the strongest method, \abt{} ranked second, and \xiang{} performs the worst.
All the baseline parsers surpass the \udpipe{} parser (row \circled{1}) in terms of the LAS and MLAS measures. Since the measure BLEX uses lemmas and \udpipe{} is much better in terms of lemmatization, it achieves higher BLEX than the baseline parsers (in fact it achieves the highest BLEX across all the compared methods).

\begin{table}[t]
\centering
\ra{1.3}
\begin{tabular}{llp{.3cm}cp{.3cm}cp{.3cm}c}  \toprule
	&  & & LAS & & MLAS & & BLEX \\ \midrule
	\circled{1} & \udpipe{} & & 76.58 & & 61.81 & & \textbf{71.39} \\ \midrule
	\circled{2} & \mate{} & & 84.96 & & 71.32 & & 63.04 \\
	\circled{3} & \abt{} (l2r) & & 83.80 & & 70.14 & & 61.82 \\
	\circled{4} & \xiang{} (l2r-rand) & & 83.39 & & 69.66 & & 61.34 \\ \midrule
	\circled{5} & \blendBl & & 86.04 & & 72.27 & & 63.83 \\
	\circled{6} & \blend & & \textbf{86.24} & & \textbf{72.46} & & 63.98 \\ \bottomrule
\end{tabular}		

\caption{Parsing accuracy (F1 scores) on the development set. The highest value in each column is bold.}
\label{tab:devparse}
\end{table}

Rows \circled{5} and \circled{6} show results of two separate blends. \blendBl{} (row \circled{5})
is an arbitrarily selected combination of 4+4+4 instances: four \mate{} instances, four \abt{} instances (two \abt{}-l2r and two \abt{}-r2l), and four \xiang{} instances (\xiang{}-l2r-rand, \xiang{}-r2l-rand, \xiang{}-l2r-embed, \xiang{}-l2r-embed). Comparing rows (\circled{2} -- \circled{4} with row \circled{5} we see that blending parsers ends with a strong boost over the baselines, which corroborates the findings of \cite{sagae-lavie:2006:HLT-NAACL06-Short,bjorkelund-EtAl:2017:CONLL}. The blended accuracy surpasses the strongest baseline parser \mate{} by more than one point.

Finally, searching for the optimal combination yields an additional small improvement of 0.2 points. 
The best combination selected by the search contains: seven instances of \mate{}, three instances of \abt{} (two \abt{}-l2r and one \abt{}-r2l) and all the instances of \xiang{}. 

\section{Subtask (B): Beyond dependency tree}
\label{sec:taskB}

The goal of Subtask (B) was to predict labeled dependency graphs and semantic labels.
The dependency graphs used in the ST were UD dependency trees extended with additional enhanced arcs. The arcs encoded shared dependents and shared governors of conjuncts. The semantic labels (e.g. Experiencer, Place, Condition) were used to annotate additional semantic meanings of tokens. 

\subsection{System description}
Our submission to the Subtask (B) followed \cite{schuster-manning:2016:LREC,candito2017enhanced} and carried out rule-based augmentation. The method consisted of two steps. First, we parsed all sentences to obtain surface dependency trees. Since the training data for Subtasks (A) and (B) was the same, we performed parsing with the same \blend{} system as described in Section~\ref{sec:description}. In the second step, we applied 12 simple rules to the predicted trees and augmented them with enhanced relations. 

The rules of the system were designed manually and guided by intuition of a Polish native speaker while analyzing gold-standard graphs from the training part of the treebank. As the enhanced relations in the treebank mostly apply to conjuncts, i.e., tokens connected with the relation ``conj'' to their heads, our rules only apply to such tokens. We define two main rules: $Head$, which predicts additional heads, and $Children$, which adds enhanced children. The remaining 10 out of the 12 rules serve as filtering steps to improve the accuracy of the $Children$ rule. \\

\begin{figure}[t]
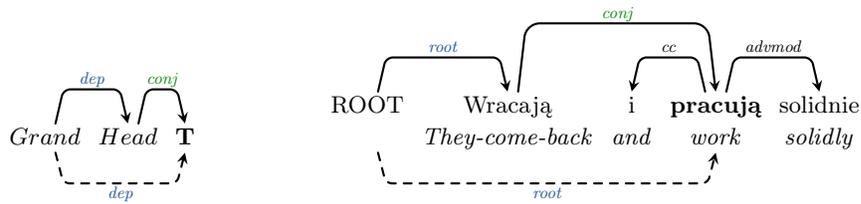

\begin{subfigure}[t]{0.33\linewidth}
\centering
\conjRule
\caption{$Head$ rule -- adds the grandparent as an additional enhanced head. Applies to all tokens with ``conj'' heads.}
\label{fig:headrule}
\end{subfigure}\hfill
\begin{subfigure}[t]{0.6\linewidth}
\centering
\conjRuleExample
\caption{Example sentence (id \texttt{train-s9826}) where the $Head$ rule introduces a correct enhanced ``root'' arc.}
\label{fig:headRuleExample}
\end{subfigure}

\label{fig:conjRule}
\caption{The $Head$ rule.} 
\end{figure}

\noindent \textbf{The $Head$ rule}
introduces enhanced arcs for all the tokens whose head is ``conj'' and connects them to their grandparents (see Figure~\ref{fig:headrule}). Figure~\ref{fig:headRuleExample} shows an example of a sentence where an enhanced arc was introduced by the $Head$ rule. The word \textit{pracują} (eng. they-work) received an additional head ROOT.

When introducing enhanced heads for ``conj'' tokens, this rule achieves an F-score of 99.40 on the gold-standard trees from the training data. \\
	
\begin{figure}[t]
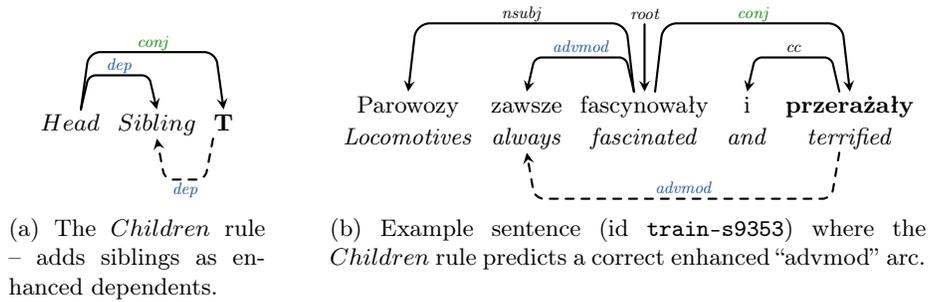

\begin{subfigure}[t]{0.28\linewidth}
\centering
\siblRule
\caption{The $Children$ rule -- adds siblings as 
enhanced dependents.}
\label{fig:childRule}
\end{subfigure}\hfill
\begin{subfigure}[t]{0.65\linewidth}
\centering
\fOneContrExample
\caption{Example sentence (id \texttt{train-s9353}) where the $Children$ rule predicts a correct enhanced ``advmod'' arc.}
\label{fig:childRuleExample}
\end{subfigure}
\caption{The $Children$ rule.}
\end{figure}

\noindent \textbf{The $Children$ rule}
adds all the siblings of a ``conj'' token as its dependents (see Figure~\ref{fig:childRule}). Figure~\ref{fig:childRuleExample} shows an example of a sentence where an enhanced arc was introduced by the $Children$ rule. The word \textit{zawsze} (eng. always)  is a sibling of the ``conj'' token \textit{przerażały} (eng. terrified) and therefore got attached to it by an ``advmod'' arc.

When introducing enhanced children of ``conj'' tokens this rule alone is too generous. On gold trees from the training data it has a perfect recall, it introduces a lot of incorrect arcs. It achieves a precision of only $21.64$, resulting in an an F-score of $35.58$. We tackled this problem by designing 10 additional filtering rules which remove some suspicious arcs. Combined with the 10 filtering rules the $Children$ rule achieves an F-score of $73.55$ on the gold trees from the training data. Below we give examples of three such rules: $labels$, $advmod_1$, $obj$. 

\paragraph{The filter $labels$} removes all the enhanced arcs with labels that are not among the ten most common ones: case, nsubj, mark, obl, advmod, amod, cop, obj, discourse:comment, advcl.

\begin{figure}
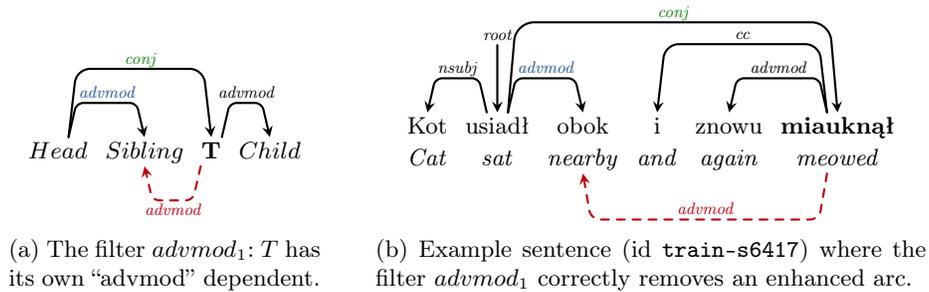

\begin{subfigure}[t]{0.34\linewidth}
\centering
\fOneGeneral
\caption{The filter $advmod_1$: $T$ has its own ``advmod'' dependent.}
\label{fig:advmod1}
\end{subfigure}\hfill
\begin{subfigure}[t]{0.6\linewidth}
\centering
\fOneExample
\caption{Example sentence (id \texttt{train-s6417}) where the filter $advmod_1$ correctly removes an enhanced arc.}
\label{fig:advmod1Example}
\end{subfigure}
\caption{The filter $advmod_1$.}
\end{figure}

\paragraph{The filter $advmod_1$} is the first of four filtering rules that remove enhanced arcs with label ``advmod''. It applies to tokens which have their own ``advmod'' basic modifiers (see Figure~\ref{fig:advmod1}). The intuition is that if the token has its own adverbial modifier then most likely the modifier of its head does not refer to it. Figure~\ref{fig:advmod1Example} shows an example of a sentence where $advmod_1$ correctly removed an arc. Since the word \textit{miaukn\k{a}\l{}} (eng. meowed) has its own adverbial modifier \textit{znowu} (eng. again) the enhanced arc to \textit{obok} (eng. nearby) was removed.

When applied to the training data, this filter removed 105 enhanced arcs with an accuracy of 93\%.

\begin{figure}[t]
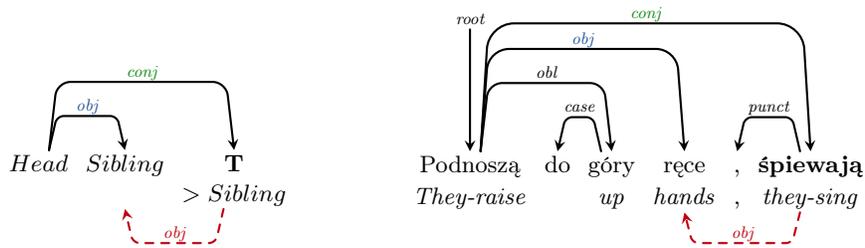

\begin{subfigure}[t]{0.33\linewidth}
\centering
\fThreeGeneral
\caption{The filter $obj$: sibling with label ``obj'' appears before $T$ in the sentence.}
\label{fig:objfilter}
\end{subfigure}\hfill
\begin{subfigure}[t]{0.6\linewidth}
\centering
\fThreeExample
\caption{Example sentence (id \texttt{train-s12456}) where the filter $obj$ correctly removes an enhanced arc.}
\label{fig:objexample}
\end{subfigure}
\caption{The filter $obj$.} 
\end{figure}

\paragraph{The filter $obj$} is the only filter which removes arcs with label ``obj''.
It applies when the enhanced ``obj'' modifier appears before the token in the sentence (see Figure~\ref{fig:objfilter}). 
The intuition is that in Polish ``obj'' modifiers tend to appear after both of the conjuncts. For example, in sentence  \textit{Podziwiali i doceniali ją też uczniowie} (id \texttt{train-s4812}; eng. Admired and appreciated her also students) the ``obj'' modifier \textit{ją} (eng. her) appears after both of \textit{Podziwiali} (eng. admired) and \textit{doceniali} (eng. appreciated) and modifies both of them. In contrast, 
Figure~\ref{fig:objexample} shows an example of a sentence where the filter $obj$ correctly removed an arc. The rule $Children$ introduced an arc from the token \textit{śpiewają} (eng. they-sing) to \textit{ręce} (eng. hands). But since the word \textit{ręce} appears before \textit{śpiewają} the arc was removed. 

When applied to the training data, this filter removed 854 enhanced arcs with an accuracy of 96\%.
   
\subsection{Evaluation of the rules}

In this section we evaluate the rules on the development set to test if they  generalize well. As a baseline we use the system without any rules, i.e., we run the evaluation script on trees without any enhanced arcs.

We start with oracle experiments and apply the rules to gold-standard trees (see Table~\ref{tab:elasgains}; Column~2). In this scenario the baseline achieves a very high accuracy of $94.23$ ELAS. Adding the $Head$ rule gives a big boost of almost 4 points, resulting in an ELAS of $98$. As expected, the pure $Children$ rule introduces too many incorrect arcs and considerably deteriorates the performance. All the consecutive filters ($labels$, $advmod_1$, $obj$) give small improvements, but together (see Table~\ref{tab:elasgains}; the final row) they not only recover the drop caused by the $Children$ rule but also improve the total accuracy by additional $0.73$ points. 

Next, we analyze the situation when enhanced arcs are introduced on automatically parsed trees. We apply the rules to outputs of two systems: the strongest parsing baseline \mate{} and the full ensemble system \blend{}. As expected, replacing gold-standard trees with a parser's predictions results in a big drop in performance: baseline accuracy decreases from $94.23$ to $80.68$ for \mate{} and $81.83$ for \blend{}. Apart from the lower starting point, the rules behave similarly to the setting with gold-standard trees: $Head$ gives a big boost, $Children$ causes a big drop in accuracy, while the 12 rules together perform better than $Head$ alone. Finally, comparing the accuracy of \mate{} and \blend{} shows that the parsing accuracy directly translates into enhanced parsing accuracy -- \blend{} surpasses \mate{} by $1.28$ in terms of LAS (cf. Table~\ref{tab:devparse}) and the advantage stays the same in terms of ELAS ($1.31$ points). 

\begin{table}[t]
\centering
\ra{1.3}
\begin{tabular}{lp{1cm}cp{1cm}cc}  \toprule
	 & & Gold & & \mate{} & \blend \\ \midrule
	No rules & & 94.23 & & 80.68 & 81.83 \\ \midrule
	+ $Head$ & & 98.00 & & 82.94 & 84.15  \\
	+ $Children$ & & 93.19 & & 78.12 & 79.25 \\
	- $labels$ & & 96.82 & & 81.26 & 82.45 \\
	- $advmod_1$ & & 96.98 & & 81.45 & 82.65 \\
	- $obj$ & & 97.37 & & 81.84 & 83.05 \\
	\midrule
	12 rules & & 98.73 & & 83.28 & 84.60  \\ \bottomrule
\end{tabular}		
\caption{Gains in enhanced parsing accuracy (ELAS) on the development set for incremental changes to the set of rules and different input trees.}
\label{tab:elasgains}
\end{table}

\section{Test Results}
\label{sec:test}

The final results on the test set are shown in Table~\ref{fig:testResults}. 
In Subtask (A) we ranked second in terms of LAS score ($83.82$) and MLAS score ($69.27$) and were behind the COMBO team by $2.29$ and $6.9$ points respectively. We achieved the third best result in terms of BLEX score due to our poor lemmatization accuracy. In Subtask (B) we ranked first with an ELAS score of $81.90$. Since we did not predict any semantic labels our SLAS score can be treated as a baseline result of running the evaluation script only on trees. 

\begin{table}[t]
\begin{subtable}[b]{0.47\linewidth}
\ra{1.3}
\centering
\begin{tabular}{lcccc} \toprule
	 & & LAS & MLAS & BLEX \\ \midrule
	COMBO & & \textbf{86.11} & \textbf{76.18} & \textbf{79.86} \\
	IMS & & 83.82 & 69.27 & 60.88 \\
	Poleval2k18 & & 77.70 & 61.21 & 70.01 \\
	Drewutnia & & 27.39 & 18.12 & 25.24 \\ \bottomrule
\end{tabular}		
\caption{Subtask (A): dependency parsing}
\label{fig:testA}
\end{subtable}\hfill
\begin{subtable}[b]{0.47\linewidth}
\ra{1.3}
\centering
\begin{tabular}{lccc} \toprule
	 & & ELAS & SLAS \\ \midrule
	IMS & & \textbf{81.90} & 65.98 \\
	COMBO & & 80.66 & \textbf{77.30} \\
	Poleval2k18 & & 66.73 & 67.84 \\ \bottomrule \\
\end{tabular}
\caption{Subtask (B): enhanced parsing}
\label{fig:testB}
\end{subtable}
\caption{Test results for all the systems participating in Task~1. The highest value in each column is bold.} 
\label{fig:testResults}
\end{table}

\section{Conclusion}
\label{sec:conclude}

We have presented the IMS contribution to \polevalFull{}.

In Subtask (A) we re-used our system from \conllstFull{}.
We confirmed our previous findings that strong preprocessing, supertags, and the use of diverse parsers for blending are important factors influencing the parsing accuracy. We extended those findings to the PolEval treebank which was a new test case for the system. The treebank differs from traditional treebanks since it is mostly built from selected sentences containing difficult syntactic constructions, instead of being sampled from some source at random.

In Subtask (B) we extended the bulky ensemble system from Subtask (A) by a set of 12 simple rules predicting enhanced arcs. We showed that a successful rule-based augmentation strongly depends on the employed parsing system. As we have demonstrated, if perfect parsing is assumed (by using gold trees), the simple rules we have developed are able to achieve an extremely high ELAS, leaving little space for further improvements. However, since the rules are not built to handle parsing errors, the parsing accuracy directly translates into performance on predicting the enhanced arcs.

\section*{Acknowledgments}
This work was supported by the Deutsche Forschungsgemeinschaft (DFG) via the SFB 732, project D8.

\bibliographystyle{./styles/splncs04}
\bibliography{references}
\end{document}